\documentclass[11pt]{article}

\usepackage[preprint]{acl}

\usepackage{times}
\usepackage{latexsym}

\usepackage[T1]{fontenc}

\usepackage[utf8]{inputenc}

\usepackage{microtype}

\usepackage{inconsolata}

\usepackage{graphicx}

\usepackage{algorithm}
\usepackage{algorithmicx}
\usepackage{algpseudocode}

\usepackage{amsmath}
\usepackage{booktabs}
\usepackage{multirow}
\usepackage{makecell}
\usepackage{tikz}
\usepackage{pgfplots}

\usepackage{colortbl}
\usepackage[table,xcdraw]{xcolor}
\usepackage{amssymb}
\usepackage{caption}
\usepackage{subcaption}
\usepackage{tabularx}

\usepackage{pifont}
\usepackage{subcaption}
\usetikzlibrary{shapes,shadows}
\usepackage[most]{tcolorbox}

\title{Mitigating Overthinking through Reasoning Shaping}

\author{
    Feifan Song$^{1,2*}$, 
    Shaohang Wei$^{1}$, 
    Bofei Gao$^{1}$, 
    Yejie Wang$^{2}$, 
    Wen Luo$^{1}$, 
    Wei Li$^{1}$\\
    \textbf{Linli Yao}$^{1}$, 
    \textbf{Weimin Xiong}$^{1}$, 
    \textbf{Liang Chen}$^{1}$, 
    \textbf{Tianyu Liu}$^{1}$\thanks{\quad Project lead.}, 
    \textbf{Houfeng Wang}$^{1}$\thanks{\quad Corresponding author.}\\
    $^{1}$State Key Laboratory of Multimedia Information Processing\\School of Computer Science, Peking University\\
    $^{2}$Moonshot AI\\
  \texttt{songff@stu.pku.edu.cn; wanghf@pku.edu.cn}
}

\newcommand\modelname{GRSP}

\begin{document}
\maketitle
\begin{abstract}
Large reasoning models~(LRMs) boosted by Reinforcement Learning from Verifier Reward~(RLVR) have shown great power in problem solving, yet they often cause overthinking: excessive, meandering reasoning that inflates computational cost.
Prior designs of penalization in RLVR manage to reduce token consumption while often harming model performance, which arises from the oversimplicity of token-level supervision. 
In this paper, we argue that the granularity of supervision plays a crucial role in balancing efficiency and accuracy, and propose \textbf{G}roup \textbf{R}elative \textbf{S}egment \textbf{P}enalization (\textbf{\modelname{}}), a step-level method to regularize reasoning. Since preliminary analyses show that reasoning segments are strongly correlated with token consumption and model performance, we design a length-aware weighting mechanism across segment clusters.
Extensive experiments demonstrate that \modelname{} achieves superior token efficiency without heavily compromising accuracy, especially the advantages with harder problems. Moreover, \modelname{} stabilizes RL training and scales effectively across model sizes.
\end{abstract}

\begin{figure*}[t]
    \centering
    \includegraphics[width=0.95\linewidth]{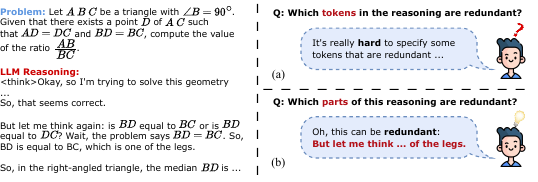} 
    \caption{Illustration of redundancy detection.
    (a)~Identifying redundant tokens is challenging, as most of them are weakly correlated with the golden answer;
    (b)~Identifying redundant steps is much easier for its clearer meaning.}
    \label{fig:head}
\end{figure*}

\section{Introduction}

Test-time Scaling with RLVR has greatly accelerated the development and adoption of Large Reasoning Models (LRMs)~\cite{team2025kimi, guo2025deepseek, yu2025dapo, zheng2025group}. During inference, LRMs typically exhibit a distinct behavior from normal post-trained LLMs: they first produce a reasoning trajectory before generating the final response.
Unlike conventional Chain-of-Thoughts~\cite{wei2022chain}, such trajectories often involve exploration and self-reflection over multiple possible solution paths, and gradually reach the final answer.
However, this pattern can also lead LRMs to overthink, or repeatedly explore and revise their paths, resulting in excessively long decoding sequences and huge computational costs.

Recent works treat this issue from the algorithmic perspective, introducing additional supervision on token efficiency within RLVR.
For example, \citet{aggarwal2025l1} penalizes samples whose output length exceeds that of the reference responses.
However, while such methods effectively reduce token consumption, they also compromise the benefits of Test-time Scaling, leading to a significant degradation in LRM performance.

Compressing the reasoning content essentially means removing redundant parts of the thought process.
For humans, it first requires the ability to be aware of the redundant content before making a decision.
When each token is considered as the candidate to remove, as illustrated in Figure~\ref{fig:head}(a), it is difficult to identify which specific tokens are unnecessary, since most tokens cannot be directly associated with the sparse verifier reward and recognized as high-value tokens.
On the other hand, arbitrarily removing tokens is not feasible either, as they are still essential to preserve the readability and coherence of the reasoning.
This motivates us to reconsider the granularity of supervision for balancing computational cost and task performance, and we find intermediate steps/segments can be a more natural unit.
As shown in Figure~\ref{fig:head}(b), humans can more easily identify a redundant step than an individual token, since each step typically carries a semantically coherent piece of thought.
Building on this insight, we propose to supervise the reasoning process at the segment level, enabling stable control of LRM reasoning behavior.

In this work, we introduce \textbf{G}roup \textbf{R}elative \textbf{S}egment \textbf{P}enalization~(\textbf{\modelname{}}), a novel method that balances computational efficiency and task performance by operating at the reasoning-step granularity. 
As a foundation, our observations on open-source LRMs reveal that the quantity of segments is positively correlated with token consumption, while they are easier to assess for redundancy, making them a more reasonable target for penalization.
Further preliminary analyses indicate a statistical relationship between the performance of LRMs and the distribution of segment lengths: stronger models tend to exhibit more balanced segment-length distributions.
It suggests a chance to mitigate performance degradation by applying length-aware penalties: penalizing segment counts within each length cluster and assigning decreasing weights for longer segments.

We conduct extensive experiments comparing \modelname{} with baselines and demonstrate its advantages in both token efficiency and accuracy. 
Notably, as task difficulty increases, together with longer reasoning, the advantages of \modelname{} become even clearer.
We also analyze the impact of the weighting mechanism.
Although it appears to contradict the intuition of encouraging shorter segments to minimize token count, our results show that it ultimately saves the cost and stabilizes RL training.
Moreover, we investigate the effect of different segmentation strategies and evaluate the scalability of our approach across model sizes.

Our work can be summarized into three aspects:\\
(1)~We first address the granularity of supervision in mitigating overthinking, proposing step-level supervision, and conducting preliminary analyses that uncover correlated features.\\
(2)~We propose \textbf{\modelname{}}, which employs length-aware weighting across segment clusters to control behavior in reasoning, mitigating performance degradation.\\
(3)~We perform comprehensive experiments to verify the broad effectiveness of segment-level penalization and weighting, and analyze its scalability on model size, offering insights for future research.

\begin{figure*}[t]
    \centering
    \includegraphics[width=1\linewidth]{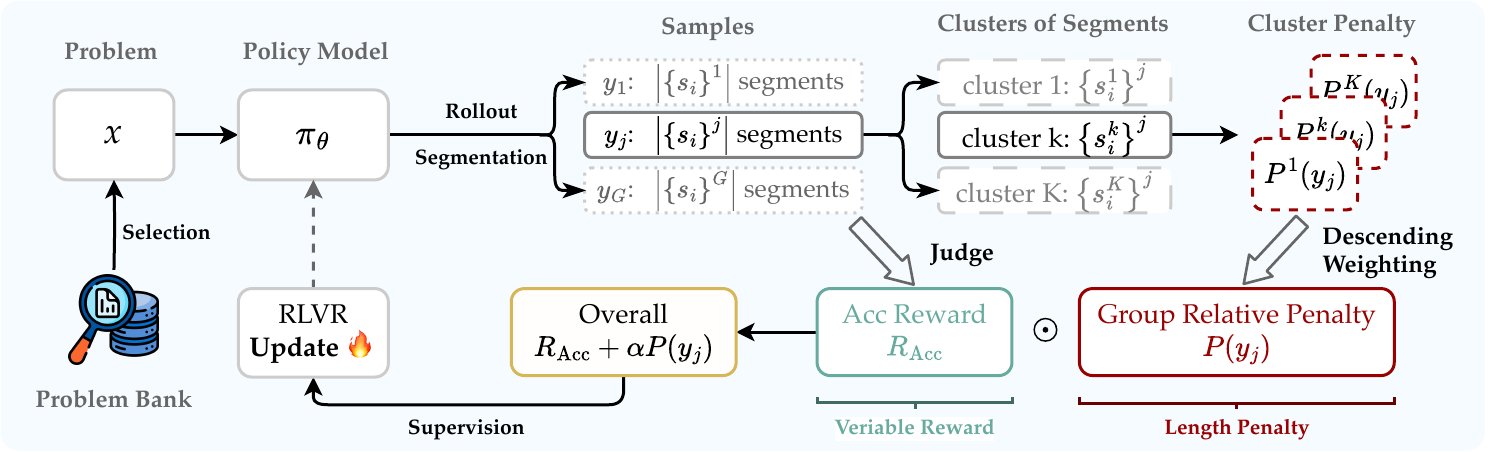} 
    \caption{The overall workflow of \modelname{}.}
    \label{fig:workflow}
\end{figure*}
\section{Preliminary}
\label{sec:preliminary}

Large reasoning models~(LRMs) improve downstream accuracy by prefixing a thinking block to the final answer. This pattern has been further scaled by recent RL work, such as Kimi-k1.5~\cite{team2025kimi} and DeepSeek-R1~\cite{guo2025deepseek}. Given the prompt $x\sim D$ and corresponding responses $\{y | y= {y_t} \sim \pi_{\theta_{\text{old}}}\}$, the goal $\mathcal{T}$ is to maximizes the expected verifiable reward,
\begin{equation}
    \mathcal{T} = \arg\max_{\pi_\theta} \mathbb{E}_{x\sim D, y\sim\pi_{\theta_{\text{old}}}(x)}R(x, y)
\end{equation}
where $\pi$ is the policy and $R$ is the accuracy signal obtained from a deterministic verifier. Prevalent algorithms include Reinforce
\begin{equation}
\begin{aligned}
    L = &\mathbb{E}_{x\sim D, Y\sim\pi_{\theta_{\text{old}}}} \frac{1}{|Y|}\sum_{y^i\in Y}\frac{1}{|y^i|}\sum_{y^i_t} \\
    & \quad \left[ A_{i,t}*\pi_{\theta}\left(y^i_t|x, y^i_{<t}\right)\right]
\end{aligned}
\end{equation}
and GRPO~\cite{shao2024deepseekmath}, which adds an importance-sampling ratio 
\begin{equation}
    r(x, y_t, y_{<t}) = \frac{\pi_\theta(y_t|x, y_{<t})}{\pi_{\theta_{\text{old}}}(y_t|x, y_{<t})}
\end{equation}
and a clip operator inherited from PPO~\cite{schulman2017proximal}. It also replaces the token-level advantage $A_{i,t}$ with a sequence-level score $A_{i}$
\begin{equation}
    A_{i} = \frac{R(x, y^i) - \mathbb{E}_{y^j\in Y}[R(x, y^j)]}{\text{std}[R(x, y^j)]}
\end{equation}
Despite its effectiveness, RLVR produces unnecessarily long thinking content. A common remedy is to augment $R$ with a token-length penalty
\begin{equation}
\label{eq:token_penalize}
    R' = R \odot P(\{y_t\})
\end{equation}
where $P$ penalises the total number of generated tokens. It shortens the output but also degrades accuracy.

\section{Methodology}
In this section, we present \modelname{}, a drop-in replacement for the token-level penalty in Eq~\ref{eq:token_penalize} in RL algorithms.
We first introduce the core segment-count penalty~($\S$~\ref{sec:segment_penalize}), then extend it with length-group weights based on our observations of reasoning cases~($\S$~\ref{sec:length_group}). Finally, we describe our segmentation methods~($\S$~\ref{sec:segmentation}). The overall workflow of \modelname{} is illustrated in Figure~\ref{fig:workflow}.

\begin{table}[t]
    \centering
    \resizebox{\linewidth}{!}{
    \begin{tabular}{l|c|c}
        \toprule
        \textbf{Model} & \textbf{\# Tokens} & \textbf{\# Segments} \\
        \midrule
        DeepSeek-R1~\cite{guo2025deepseek} & 8544.70 & 135.75 \\
        QwQ-32B~\cite{yang2024qwen2} & 12003.31 & 216.52 \\
        DS-Qwen-Distill-32B & 7464.53 & 105.51 \\
        DS-Qwen-Distill-14B & 6539.29 & 94.73 \\
        \bottomrule
    \end{tabular}
    }
    \caption{Statistics of open-source LRMs on token consumption and reasoning segment production.}
    \label{tab:preliminary}
\end{table}

\begin{figure*}[t]
    \centering
    \includegraphics[width=1\linewidth]{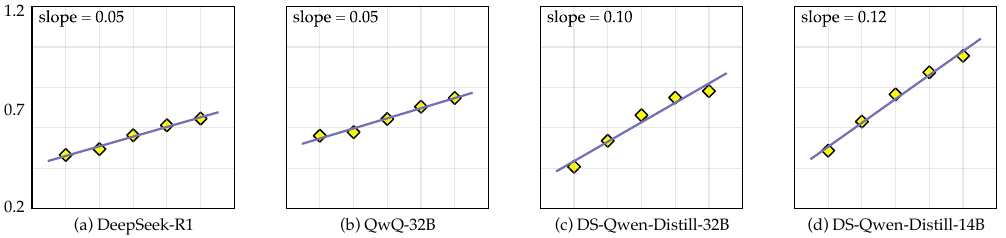} 
    \caption{The ratio of segment counts across each cluster (correct vs. wrong). Longer segments are generally more prevalent in correct cases, and stronger models (a, b) exhibit flatter slopes compared to weaker ones (c, d).}
    \label{fig:ref_models}
\end{figure*}

\subsection{Penalization on Segments}
\label{sec:segment_penalize}
Mitigating overthinking essentially means compressing the thinking content. 
As shown in Figure~\ref{fig:head}, identifying redundant tokens is often ambiguous.
However, distinct steps, i.e., trajectory segments within the reasoning process, can serve as a natural higher-level unit, as also adopted by \citet{guo2025segment}.
Moreover, our investigation of several open-source LRMs reveals a clear positive correlation between the quantity of segments and total token consumption (see Table~\ref{tab:preliminary}).
Therefore, by decomposing the thinking content into segments $\{s_i\}$, we can indirectly reduce token usage by discouraging unnecessary steps.

To be specific, for a single prompt $x$ and a corresponding response group $Y = \{y^j\}$ sampled from $\pi_{\theta_{\text{old}}}$, let ${s_i}^j$ denote the trajectory segments in response $y^j$.
We then compute the z-score of the segment count inside the group $Y$ for each $y^j\in Y$, and treat its negative value as the penalty, which requires no task-specific threshold:
\begin{equation}
\label{eq:z-score}
    P(y^j) = - \frac{|\{s_i\}^j| - \mathbb{E}_{y\in Y}[|\{s_i\}|]}{\text{std}[|\{s_i\}|]}
\end{equation}

\subsection{Group Relative Penalization}
\label{sec:length_group}

Simply changing the granularity of penalization from tokens to trajectory segments does not solve the problem in §\ref{sec:preliminary}, because the performance gain of an LRM over an ordinary post-trained model comes directly from \textbf{generated-token scaling}, and any mechanism that shortens the reasoning consequently risks a sharp drop in performance.

To be specific in the RL setting, continued improvement on downstream performance is usually obtained by stable training or more optimization steps~\cite{yu2025dapo,zheng2025group}.
However, introducing a length penalty can destabilise it: there is a phenomenon that once the response length exceeds a certain point, the penalty dominates and gradually the training collapses with task accuracy drifting downward. Hence, how to discourage over-length thinking while keeping the training process stable becomes the central question.

We investigate the above LRMs again for statistical signatures that correlate with these two requirements.  
We further hypothesize that model performance is correlated with the extent of training, i.e., stronger models tend to undergo more training, which in turn suggests greater training stability.
In detail, using the RL problems in \S~\ref{sec:experimental_setup}, we roll out a batch of responses for each model. For these responses, we segment the thinking content and group the segments into five clusters by length, where segments longer than 300 tokens are excluded, as they rarely occur.
Since this experiment is related to model performance, we apply the above procedure separately to passed and failed cases, and compute the average number of segments in each cluster for the two sides.
Intuitively, we find that failed cases tend to contain more segments across most clusters. We then calculate, for each length cluster, the ratio of the average number of segments from the passed side to that from the failed side. The results are shown in Figure~\ref{fig:ref_models}.

This experiment leads to several findings:\\
(1) It is a general trend that failed cases contain more segments than passed ones.\\
(2) Errors of LRMs in problem solving can be associated with the presence of relatively shorter segments in the thinking content. In detail, the difference between passed and failed cases is larger in shorter-length clusters (with lower ratios in Figure~\ref{fig:ref_models}), while the difference decreases in longer-length clusters (with higher ratios).
It is a shared pattern for all examined LRMs.\\
(3) A potential link between model performance (+ training stability) and a more balanced distribution of segments can exist. Stronger models show a smaller variation in this ratio across clusters, which is reflected in a smaller slope of the linear fit.

Based on these observations, we propose stabilizing RL training by explicitly shaping the distribution of segments through segment-level penalties, which we term \textbf{Reasoning Shaping}. Concretely, following the procedure above, we first split the thinking content into segments and cluster them by length. For each cluster $k$, we compute a relative penalty using a z-score normalization:
\begin{equation}
P^k(y^j) = - \frac{|{s^k_i}^j| - \mathbb{E}_{y\in Y}[|{s^k_i}|]}{\text{std}[|{s^k_i}|]}
\end{equation}
The overall penalty is then obtained by weighting across clusters:
\begin{equation}
P(y^j) = \sum_{k} w^k P^k(y^j)
\end{equation}
Following the findings above, we penalize short segments more heavily, while applying weaker penalties to longer ones. To achieve this, we assign \textbf{descending} weights from shorter to longer clusters. The final reward is then given by:
\begin{equation}
R' = R + \alpha P(y^j)
\end{equation}

\subsection{Segmentation}
\label{sec:segmentation}

We design two mechanisms for segmentation.
The first is keyword-based matching, similar to \citet{guo2025segment}. We collect a list of keywords from typical reasoning cases to identify potential segment boundaries. It is used by default for conciseness and high computational efficiency, while its applicability is limited to specific languages.

The second mechanism is token log-probability matching. Following observations in \citet{li2024cascade, song2025well, wang2025beyond}, segment boundaries often correspond to local minima in token log-probabilities. This is because segment transitions usually admit more candidate continuations, leading to lower confidence at the beginning of a new segment. Based on it, we locate boundaries by identifying these local minima. We implement and test it in $\S$~\ref{sec:logits_seg}.

\begin{table*}[ht]
    \centering
    \resizebox{\linewidth}{!}{
    \begin{tabular}{l|cc|cc|cc|cc}
        \toprule
         \multirow{2}{*}{Model / Method} & \multicolumn{2}{c|}{\textbf{MATH 500}} & \multicolumn{2}{c|}{\textbf{AIMO Prize 1}} & \multicolumn{2}{c|}{\textbf{Omni-MATH 500}} & \multicolumn{2}{c}{\textbf{Overall}} \\
         & Acc. & Avg Len. & Acc. & Avg Len. & Acc. & Avg Len. & Acc. & Avg Len. \\
         \midrule
         \multicolumn{9}{c}{\textbf{Open-source Models}} \\
         \midrule
         DeepSeek-R1 & 96.60 & 2428 & 91.25 & 4704 & 70.80 & 7456 & 84.26 & 4924 \\
         QwQ-32B & 98.00 & 4260 & 95.00 & 9222 & 71.20 & 12125 & 85.37 & 8268 \\
         DS-Qwen-Distill-32B & 96.40 & 2594 & 82.50 & 6650 & 61.80 & 8997 & 79.35 & 5859 \\
         DS-Qwen-Distill-14B & 93.80 & 2538 & 80.00 & 6504 & 63.60 & 9091 & 78.80 & 6075 \\
         Qwen-2.5-14B-it & 54.40 & 488 & 10.00 & 1091 & 16.60 & 918 & 33.61 & 811 \\
         Qwen-2.5-14B-it$^*$ & 84.80 & 1460 & 53.75 & 2579 & 39.80 & 2159 & 61.67 & 2116 \\
         \midrule
         \multicolumn{9}{c}{\textbf{Training Qwen-2.5-14B-it$^*$}} \\
         \midrule
         Reinforce & 87.20 & 1887 & 53.75 & 3568 & 44.20 & 5131 & 64.81 & 3513 \\
         \quad + LCPO & \textbf{86.40} & 2222 & 57.50 & 3771 & 42.40 & 7994 & 63.89 & 5009 \\
         \quad + O1-Pruner & 85.20 & \textbf{1738} & \textbf{60.00} & 3416 & 40.40 & 5226 & 62.59 & \textbf{3477} \\
         \quad + \modelname{} & 85.40 & 2128 & 55.00 & \textbf{3215} & \textbf{45.60} & \textbf{4866} & \textbf{64.72} & \textbf{3477} \\
         \midrule
         GRPO & 85.20 & 2131 & 60.00 & 2648 & 45.40 & 5315 & 64.91 & 3643 \\
         \quad + LCPO & 86.00 & 1919 & 52.50 & 3597 & 43.40 & 5855 & 63.80 & 3866 \\
         \quad + O1-Pruner & 85.00 & \textbf{1706} & \textbf{61.25} & 3299 & 40.60 & 5497 & 62.69 & 3579 \\
         \quad + \modelname{} & \textbf{86.20} & 2054 & 60.00 & \textbf{2826} & \textbf{43.80} & \textbf{4897} & \textbf{64.63} & \textbf{3427} \\
        \bottomrule
    \end{tabular}
    }
    \caption{Results of different models/methods across different benchmarks. The \textbf{higher} accuracy (Acc.) and \textbf{lower} average length of responses (Avg Len.) are expected. Models labeled by $^*$ have been SFT-tuned.}
    \label{tab:main_results}
\end{table*}

\section{Experiment}

\subsection{Experimental Setup}
\label{sec:experimental_setup}
The training pipeline consists of two post-training stages: supervised fine-tuning (SFT) followed by RLVR.
For SFT, problems are collected from NuminaMATH~\cite{numina_math_datasets}, while completions of reasoning and response are O1-mini patterned examples, following the prompt-engineering procedure of \citet{gao2025towards}.
We utilize these data to warm up the base LLM and to establish a strong initialization for subsequent RL.
For the RL stage, we use more challenging problems sampled from Omni-MATH and AIME, which tend to induce longer reasoning trajectories than NuminaMATH, and therefore better expose test-time scaling effects. The data volume and implementation detail are summarized in Appendix~\ref{appendix:data} and \ref{appendix:imple}.

Moreover, we set Qwen-2.5-14B-it as the starting checkpoint, and mark it with $^*$ after SFT. During RL, models are trained with Reinforce and GRPO, with the addition of a length penalty.

\subsection{Evaluation}
For evaluation, we construct three test sets with increasing difficulty: 500 problems from MATH 500, 10 challenging problems from AIMO Prize-1, and 500 difficult problems from Omni-MATH 500, which enables us to compute general scores and analyze how task difficulty influences model behavior and token efficiency.
Hence, we report two metrics: task performance measured by accuracy and token efficiency measured by the average number of decoding tokens per example. We compare \modelname{} with two baselines:\\
(1) LCPO~\cite{aggarwal2025l1} uses the ratio between the generated response length and the reference one as a weighting factor for the verifiable reward, reducing the sparsity of reward distribution and introducing a direct correlation between reward and token usage.\\
(2) O1-Pruner~\cite{luo2025o1} computes a ratio factor similar to LCPO, while adding it as an auxiliary reward term to the verifiable reward. 

\subsection{Main Results}
In this section, we report results on the above benchmarks.
We first evaluate four open-source LRMs introduced in \S~\ref{sec:segment_penalize}.
Although all of them exhibit strong performance, there remains a clear gap between DeepSeek-R1/QwQ-32B and the two distill models.
Notably, with 671B parameters, DeepSeek-R1 demonstrates remarkable token efficiency, suggesting that stronger base capability can contribute to more concise reasoning.
In addition, the fine-tuned Qwen-2.5-14B-it$^*$ achieves a substantial improvement over Qwen-2.5-14B-it, highlighting the effectiveness of test-time scaling.

\begin{figure*}[t]
    \centering
    \includegraphics[width=1\linewidth]{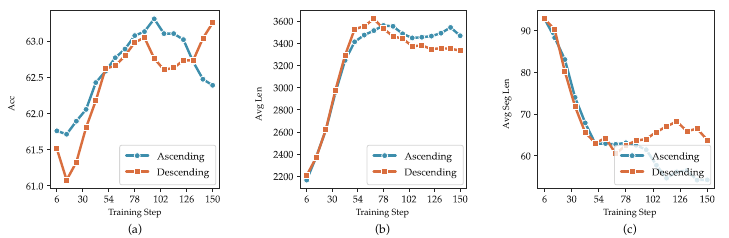} 
    \caption{Comparison of two weighting strategies.
    (a)~Accuracy over training steps;
    (b)~Average response length over training steps;
    (c)~Average segment length over training steps.}
    \label{fig:trend}
\end{figure*}
Overall, \modelname{} achieves the \textbf{best} scores on both task performance and token efficiency under both GRPO and REINFORCE frameworks.
Notably, on the most challenging Omni-MATH 500, \modelname{} achieves the most significant reduction in token usage while maintaining the highest accuracy among all baselines, even matching or surpassing the naive RL version (Reinforce + \modelname{}).
Although LCPO also performs competitively, it fails to deliver noticeable gains in token efficiency.

We further analyze how token length correlates with task difficulty.
For all methods, token overhead increases as problems become more challenging, indicating that models rely on \emph{length scaling} to tackle complex reasoning tasks.
RL accentuates this trend, as the token growth on Omni-MATH is much larger than on easier benchmarks, and \modelname{} primarily reduces token usage on such problems, where overthinking is most likely to occur, while preserving task performance.

We also observe distinct segmentation patterns across different methods, which may strongly correlate with their results. Using the keyword-based segmentation on all models trained with Reinforce, we find that \modelname{} produces an average of 21.07 segments, notably fewer than 26.66 from the model trained without additional penalty. 
This confirms that \modelname{} effectively regulates the number of reasoning segments and thus reduces token overhead.
In contrast, the two baselines yield substantially more segments of 51.97 and 142.12, respectively, and the largest part is short segments, as the ratio of segments from the cluster of $k=1$ is 79.17\% and 91.36\% for O1-pruner and LCPO, while that for \modelname{} is 62.61\%.
We attribute this to the ambiguity in supervising token length: it disturbs the optimization on verifiable reward, making the model resort to rapidly iterating short reasoning steps to maintain performance.
Such settings not only lead to lower accuracy but also have a larger likelihood of overthinking, as extremely long outputs are observed in LCPO.
We provide further analyses of this pattern in \S~\ref{sec:weighting}.

\subsection{Ablation on the Weighting Mechanism}
\label{sec:weighting}

In this section, we investigate the effect of the proposed weighting mechanism across segment clusters.
By default, the penalty weight decreases with the cluster index $k$, that is, longer segments receive smaller penalties~(\textbf{Descending}), which encourages potentially deeper reasoning within each segment while discouraging the overproduction of short segments.
As a contrastive setting, we reverse the order to make the weights increase with $k$, defined as $w^k = (k-1) \times 0.05 + 1$~(\textbf{Ascending}).
We train both configurations under REINFORCE on Qwen-2.5-14B-it$^*$ and track the evolution of accuracy~(Acc), average token quantity of responses~(Avg Len), and average segment length~(Avg Seg Len), smoothed and visualized in Figure~\ref{fig:trend}.

The results reveal that test-time scaling emerges under both settings: as training progresses, the average token quantity grows alongside task performance.
However, the Ascending configuration exhibits a much steeper rise in both metrics, reaching a peak earlier but soon suffering a sharp accuracy collapse, indicating training instability.
In contrast, the Descending configuration shows steadier improvement, with accuracy increasing more smoothly over time.
However, after reaching the peak length, the Ascending model fails to compress thinking content effectively. Considering its high weight on the short segment clusters, it resembles the degenerate feature observed in LCPO, where the model excessively expands reasoning without meaningful gains.

The segment-level patterns in Figure~\ref{fig:trend}~(c) further support this observation.
At the beginning, RLVR generally drives models to produce shorter segments to activate length scaling, and both configurations behave similarly. However, as training continues, the Ascending weights push the model to rapidly over-optimize for shorter segments, which is a turning point that coincides with its accuracy collapse.
Conversely, the Descending configuration gradually shifts toward generating longer segments, i.e., increasing the proportion of long segments.
Interestingly, this adjustment leads to an overall \textbf{shorter} average response length, suggesting that the model learns a more efficient reasoning strategy: thinking more thoroughly within each segment, thus requiring fewer total steps to reach the correct answer.

These findings also shed light on the dynamics between the verifiable reward and the length penalty.
During the early stage, the verifiable reward dominates, and both configurations exhibit similar trends across all three metrics.
In the mid-stage, both encounter a transition point where accuracy begins to drop a bit. In the near time, length decreases, indicating a shift in optimization focus from purely correctness to brevity, which in turn risks destabilizing training.
The Descending configuration, however, escapes this collapse by increasing the proportion of long segments, allowing accuracy and efficiency to improve jointly.
This confirms that accuracy and length optimization are not a zero-sum game; with a proper pattern, \modelname{} achieves a stable balance between concise reasoning and task effectiveness. Hence, we conclude that descending weighting can be a more stable and interpretable optimization signal.

\subsection{Scaling of Model Capacity}
\begin{figure}[t]
    \centering
    \includegraphics[width=1\linewidth]{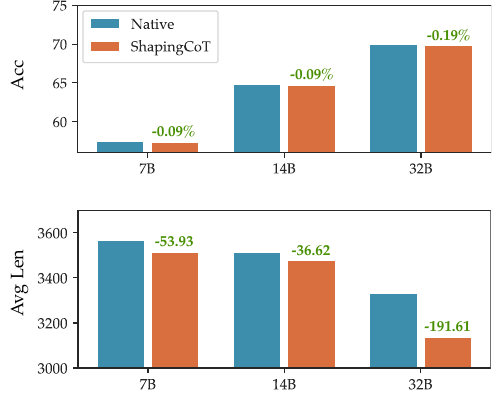} 
    \caption{Effect of \modelname{} across models of varying capacities, comparing changes in accuracy and average response length.}
    \label{fig:model_scale}
\end{figure}

\begin{figure*}[t]
    \centering
    \includegraphics[width=1\linewidth]{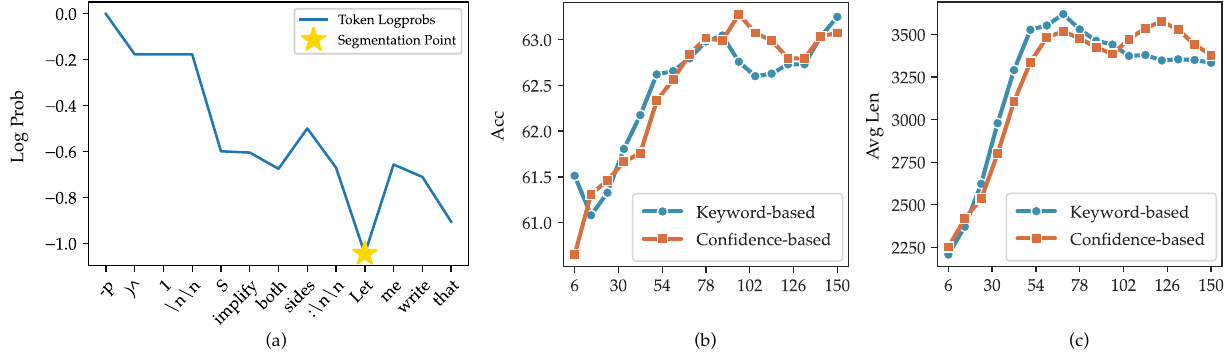} 
    \caption{(a)~Illustration of the log-probability trend around a segmentation point;
    (b)~Comparison of accuracy between keyword-based and confidence-based segmentation;
    (c)~Comparison of average response length.}
    \label{fig:logits_seg}
\end{figure*}
Results on open-source LRMs from Table~\ref{tab:main_results} suggest that the capacity of base models plays a critical role in task performance and token efficiency.
For example, DeepSeek-R1, the largest model in our evaluation, achieves near the top accuracy and token efficiency, while DS-Qwen-Distill-32B also requires fewer tokens than its 14B counterpart.
This naturally raises the question of how model capacity interacts with \modelname{} during RL training.

To investigate it, we conduct experiments on three models from the Qwen-2.5 family: 7B-it, 14B-it, and 32B-it.
Each model is first warmed up with the same SFT dataset, followed by RL training under two settings: standard Reinforce and Reinforce + \modelname{}, with all other hyperparameters held constant.
We still track accuracy and average response length to test how the effect of \modelname{} varies across model scales (Figure~\ref{fig:model_scale}). Our results reveal two clear trends:\\
(1)~Larger models are inherently more efficient and accurate even in RL, where the accuracy improves steadily with model size, while token consumption decreases under the same training setup.\\
(2)~\modelname{} consistently improves efficiency across all scales, with minimal impact on accuracy.
While accuracy remains nearly unchanged, the reduction in token usage can grow. In particular, the 32B model can complete tasks with significantly fewer tokens compared to smaller models, reflecting its greater capacity to leverage reasoning compression.

\subsection{Confidence-based Segmentation}
\label{sec:logits_seg}

In $\S$~\ref{sec:segmentation}, we introduce another segmentation strategy based on the model’s confidence distribution. Unlike keyword matching, it does not rely on manually collected keywords yet achieves similarly strong performance.

Transitions between reasoning segments often correspond to local minima in the model’s log probabilities. However, low logprob values can also occur in other cases, such as generating digits or single characters. Hence, after identifying positions where the smoothed logprob falls below a threshold $\gamma$, we further filter them based on token length and whether they correspond to local minima. Figure~\ref{fig:logits_seg}~(a) illustrates an example where a local minimum coincides with the start of a new reasoning segment.

We conduct experiments using the confidence-based segmentation under the Reinforce+\modelname{} framework on Qwen-2.5-14B$^*$, and compare it with the proposed keyword-based segmentation in terms of accuracy and response length. As shown in Figure~\ref{fig:logits_seg}~(b, c), the two curves exhibit similar trends. Notably, the confidence-based segmentation shows a rise–fall–rise pattern in both accuracy and length, whereas the keyword-based segmentation displays a steady decline in length without recovery. These patterns are consistent with our findings in $\S$~\ref{sec:weighting}: the verifiable signal and the length penalty occasionally compete, leading to temporary degradation in one metric that later recovers, instead of collapsing entirely. Ultimately, the confidence-based segmentation achieves a higher accuracy–length pair (64.91, 3415) than the keyword-based segmentation (64.72, 3477), confirming the effectiveness of this design.

\section{Related Work}
\subsection{Test-time Scaling of LLMs}
Test-time scaling has proved effective at boosting LLM performance on complex question answering, mathematical problem solving, and code generation~\cite{wu2024inference, wang2022self, wei2022chain, guo2025deepseek}. One line of such work is Monte Carlo Tree Search or tree-structured prompting~\cite{wu2024inference, yao2023tree}. However, it involves trivial human-crafted engineering that hinders scaling up. Another line is Reinforcement Learning with Verifiable Reward~(RLVR)~\cite{team2025kimi, guo2025deepseek,muennighoff2025s1,ye2025limo} which encourages exploration of long and complex reasoning paths via simple binary signals provided by verifiers.

\subsection{Efficient Long Chain-of-Thought LLMs}
Despite the remarkable effectiveness of the long reasoning patterns, they suffers from substantial computational overhead, especially for challenging user inputs where the model tends to repeatedly deliberate, named \textbf{overthinking}.
To mitigate such phenomena, recent work aims at shortening reasoning trajectories to reduce computation, while preserving task performance, e.g., accuracy~\cite{sui2025stop}. 
Lightweight approaches include designing control prompt, such as indicating a token budget or specifying reasoning granularity~\cite{DBLP:conf/acl/HanWFZM025,DBLP:journals/corr/abs-2502-18600,DBLP:journals/corr/abs-2503-05179}, or intervening in the decoding process~\cite{liu2025adaptivestep,liao2025reward,ding2025dynamic,fu2024efficiently,huang2025efficient,taubenfeld2025confidence,wang2025sampling}.
However, they do not essentially modify the model’s reasoning patterns, which may limit their robustness and ultimate performance.

A fundamental solution is to introduce supervision on reasoning efficiency during fine-tuning~\cite{xia2025tokenskip,zhang2025lightthinker,arora2025training,aggarwal2025l1,luo2025o1,team2025kimi}, especially in RLVR settings where both positive and negative rewards can effectively guide model behavior.
Nevertheless, existing studies mostly impose penalties on tokens, while the correlation between token-level penalties and the final verifiable rewards remains weak, leading to limited compression effectiveness or even degradation in accuracy.

\section{Conclusion}
To mitigate the overthinking of Large Reasoning Models (LRMs), we propose Group Relative Segment Penalization (\modelname{}), a step-level method that regularizes reasoning in RLVR. 
By supervising at the granularity of reasoning segments and applying length-aware weighting, \modelname{} effectively mitigates performance degradation while still increasing efficiency. 
We conduct extensive experiments to verify that \modelname{} has the advantages in the above two aspects, while also improving training stability and scaling across model sizes. We hope our study could inspire future research on the granularity selection of supervision during the design of RLVR algorithms for LRMs.

\section*{Limitations}
We also experimented with warm-up data constructed from other patterns, such as those derived from DeepSeek-R1. However, we found that such data tends to make the model generate overly long responses even before reinforcement learning begins. As a result, it becomes difficult to observe clear scaling effects on response length.

To further validate the compression of token usage, it would be ideal to continue training from our current checkpoint using reinforcement learning. Nevertheless, this requires substantial computational resources beyond our current budget. Despite these limitations, our experiments demonstrate the robustness of the proposed method and its potential for continued efficient scaling. We hope future work will further explore these directions.


\onecolumn
\appendix

\section{Statistics of Utilized Data}
\label{appendix:data}
Please refer to Table~\ref{tab:data}.

\begin{table}[ht]
    \centering
    \resizebox{0.5\linewidth}{!}{
    \begin{tabular}{l|l|c}
        \toprule
        \textbf{Stage} & \textbf{Data Source} & \textbf{\#} \\
        \midrule
        \multirow{1}{*}{SFT} & NuminaMath-1.5~\citep{numina_math_datasets} & 27621 \\
        \midrule
        \multirow{2}{*}{RL} & AIME & 800 \\
        & Omni-MATH~\citep{gao2025omnimath} & 2400 \\
        \midrule
        \multirow{3}{*}{Eval} & MATH 500~\citep{lightman2024lets} & 500 \\
        & AIMO Prize 1 & 10 * 8 \\
        & Omni-MATH 500~\citep{gao2025omnimath} & 500 \\
        \bottomrule
    \end{tabular}
    }
    \caption{Statistics of data utilized for SFT, RL, and evaluation, respectively.}
    \label{tab:data}
\end{table}

\section{Implementation Details}
\label{appendix:imple}
We set the maximum sequence length to 30K to accommodate long reasoning trajectories.
Each RL training run consists of 150 steps, with a rollout performed every 3 steps.
The learning rate is set to 2e-6.
We randomly sample 1024 problems for each rollout iteration, and for each problem, we conduct 10 rollouts per iteration with temperature as 1.0, while temperature for evaluation is 0.6.
Dynamic sampling~\cite{yu2025dapo} is applied, so the actual batch size varies across iterations.
For \modelname{}, the descending weights for each length cluster $w^k$ are determined by $w^k = (K - k) \times 0.05 + 1$, where the maximum cluster index $K$ is 5. 
The balancing coefficient $\alpha$ is set to 5e-3 for keyword-based segmentation and 2.5e-3 for confidence-based segmentation.
We report the results with the highest task performance.

\section{GRPO Ablation on the Weighting Mechanism}
\begin{figure*}[ht]
    \centering
    \includegraphics[width=1\linewidth]{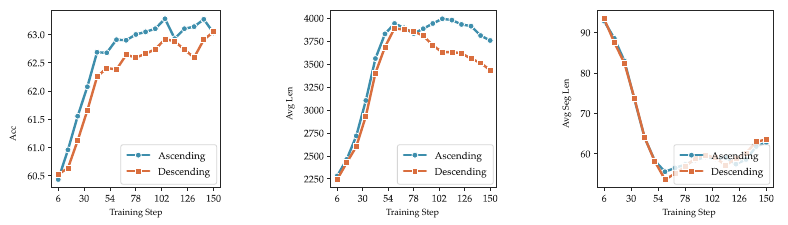} 
    \caption{Comparison of two weighting strategies for GRPO.
    (a)~Accuracy over training steps;
    (b)~Average response length over training steps;
    (c)~Average segment length over training steps.}
    \label{fig:trend_grpo}
\end{figure*}

Figure~\ref{fig:trend_grpo} presents the ablation results of the weighting mechanism on GRPO.
A similar trend to Reinforce in Figure~\ref{fig:trend} can be observed, where the Ascending weighting improves accuracy more rapidly while the Descending weighting achieves stronger compression and maintains high accuracy.
In addition, we find that GRPO training is generally more stable than Reinforce across all methods, so the Ascending setting does not collapse in the middle stage and excessively encourages short segments as in Reinforce training.
Nevertheless, the Descending weighting still produces longer reasoning segments as expected.

\section{Confidence-based Segmentation on GRPO}
\begin{figure*}[t]
    \centering
    \includegraphics[width=0.75\linewidth]{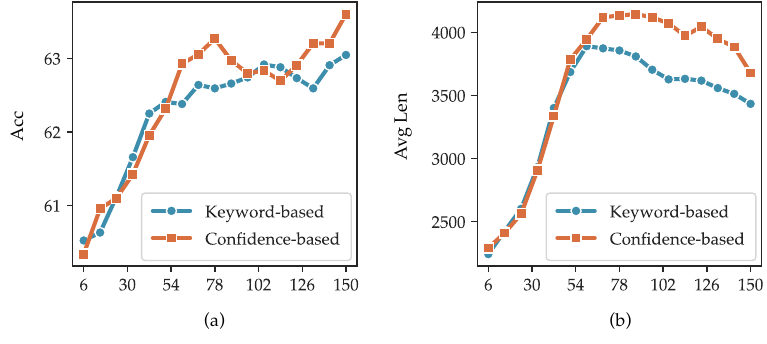} 
    \caption{GRPO results. (a)~Comparison of accuracy between keyword-based and confidence-based segmentation;
    (b)~Comparison of average response length.}
    \label{fig:logits_seg_grpo}
\end{figure*}

Figure~\ref{fig:logits_seg_grpo} illustrates the effect of confidence-based segmentation on GRPO, compared to the default keyword-based segmentation. It acquires higher accuracy in Figure~\ref{fig:logits_seg_grpo} (a). In fact, it reaches a more surprising performance of (64.81, 3405) for accuracy and average response length than (64.63, 3427) of GRPO + \modelname{} + keyword-based segmentation. Note that around the last 20 steps actually witness a significant drop in response length, enabling the model to be both powerful and efficient in reasoning, which cannot be shown in Figure~\ref{fig:logits_seg_grpo} (b) due to the curve smoothness.

\section{Prompt Template}
\begin{figure*}[h]
  \centering
  \resizebox{1\textwidth}{!}{
  \begin{tcolorbox}
    <|im\_start|>system

    You are a helpful assistant. You should first try a long-text process of thinking and reflection to handle the problem in the mind before each responding to the user. The thinking process are enclosed within <think>{\textbackslash}n and </think>{\textbackslash}n{\textbackslash}n tags, respectively, i.e., <think>
    
    [thinking process here]</think>
    
    [final answer here].
    
    <|im\_end|>
    
    <|im\_start|>user
    
    {The problem text}
    
    Please reason carefully step by step, reflect thoroughly, and put the final answer in {\textbackslash}boxed\{\{\}\}.
    
    <|im\_end|>

  \end{tcolorbox}
  }
  \caption{
  The prompt template used for SFT and RL.}
  \label{fig:prompt_1}
\end{figure*}


\end{document}